\documentclass[runningheads]{llncs}

\usepackage{eccvabbrv}

\usepackage{graphicx}
\usepackage{booktabs}

\usepackage[accsupp]{axessibility}  % Improves PDF readability for those with disabilities.

\usepackage{eccv}
\usepackage[pagebackref,breaklinks,colorlinks]{hyperref}

\usepackage{orcidlink}

\usepackage{multirow}
\usepackage{colortbl}
\definecolor{first}{rgb}{1, 0.7, 0.7}
\definecolor{second}{rgb}{1, 0.85, 0.7}
\definecolor{third}{rgb}{1, 1, 0.7}
\usepackage{adjustbox}

\begin{document}

\title{Gaussian Time Machine: A Real-Time Rendering Methodology for Time-Variant Appearances} 

\titlerunning{Gaussian Time Machine}

\author{Licheng Shen\inst{1} \and
Ho Ngai Chow\inst{1} \and
Lingyun Wang\inst{2} \and
Tong Zhang\inst{2} \and
Mengqiu Wang\inst{2, 3} \and
Yuxing Han\inst{1}
}

\authorrunning{L.~Shen et al.}

\institute{Tsinghua Shenzhen International Graduate School, Tsinghua University\and
Zero-Zero Lab \and
Zhejiang University\\
\email{slc23@mails.tsinghua.edu.cn, yuxinghan@sz.tsinghua.edu.cn}}

\maketitle

\begin{abstract}
  Recent advancements in neural rendering techniques have significantly enhanced the fidelity of 3D reconstruction. Notably, the emergence of 3D Gaussian Splatting (3DGS) has marked a significant milestone by adopting a discrete scene representation, facilitating efficient training and real-time rendering. Several studies have successfully extended the real-time rendering capability of 3DGS to dynamic scenes. However, a challenge arises when training images are captured under vastly differing weather and lighting conditions. This scenario poses a challenge for 3DGS and its variants in achieving accurate reconstructions. Although NeRF-based methods (NeRF-W, CLNeRF) have shown promise in handling such challenging conditions, their computational demands hinder real-time rendering capabilities. In this paper, we present Gaussian Time Machine (GTM) which models the time-dependent attributes of Gaussian primitives with discrete time embedding vectors decoded by a lightweight Multi-Layer-Perceptron(MLP). By adjusting the opacity of Gaussian primitives, we can reconstruct visibility changes of objects. We further propose a decomposed color model for improved geometric consistency. GTM achieved state-of-the-art rendering fidelity on 3 datasets and is 100 times faster than NeRF-based counterparts in rendering. Moreover, GTM successfully disentangles the appearance changes and renders smooth appearance interpolation.
\end{abstract}

\section{Introduction}
\label{sec:intro} 

\begin{figure}
    \centering
    \includegraphics[width=\linewidth, trim={0.4cm, 6cm, 0.4cm, 6cm}, clip]{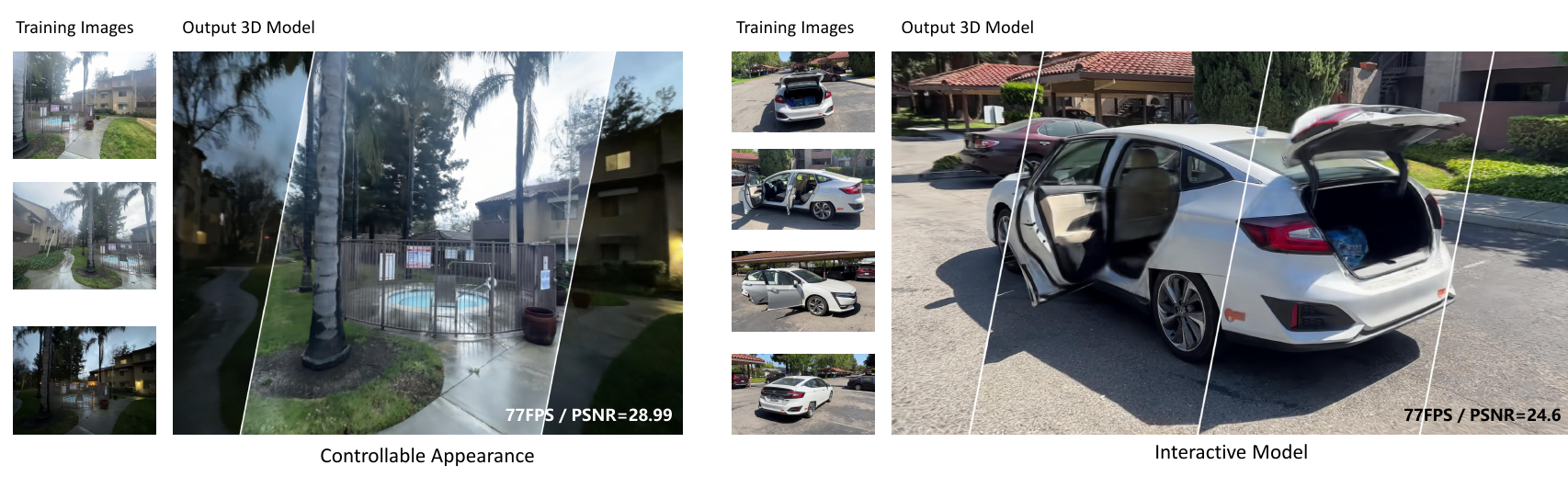}
    \caption{Gaussian Time Machine (GTM) addresses the challenge of reconstructing time-variant 3D scenes with complicated appearance changes using training images taken at discrete moments. We apply a lightweight neural time encoder, empowering 3DGS to disentangle scene variations. GTM has combined strengths of high quality, real-time speed, and flexibility in handling both lighting changes and transformations of objects, only with a single encoder and 3D Gaussian point cloud.}
    \label{fig:teaser}
\end{figure}

Data-driven methodologies have emerged as the forefront approach in novel view synthesis, primarily due to their ability to produce photo-realistic renderings. Recently, 3D Gaussian Splatting (3DGS)\cite{kerbl20233d} has proposed an innovative scene representation, enhancing training speed, rendering efficiency and quality. Notably, several studies have extended the application of 3DGS to dynamic scenes while maintaining real-time rendering speed, showcasing its adaptability to varying environmental conditions and rendering requirements.

However, a significant challenge emerges when training images are captured at discrete times, exhibiting varying weather and lighting conditions. While some NeRF-based\cite{Martin-Brualla_2021_CVPR, Cai_2023_ICCV, tancik2022block} methods demonstrate the capability to disentangle appearance changes, their computational costs limit the rendering speed. On the other hand, although 3DGS and its variants\cite{4dgs, yang2023deformable3dgs} excel in capturing detailed scene dynamics within a limited range or short timeframe, reconstructing scenes with varying appearances remains a challenging tasks. A naive approach is to train separate models for each appearance, which is inefficient because of extra training cost.

We propose Gaussian Time Machine(GTM), a real-time rendering method for scenes with vastly varying appearances. We use lightweight neural networks to predict the attributes of Gaussian primitives. Then we add time embedding vectors to the inputs of the opacity and color encoder. We further decompose rendering color into a static term and a dynamic term. With such a design, scenes with varying appearance can be reconstructed efficiently while maintaining geometry consistency. 

GTM is tested on public datasets of real-world scenes. Experimental results show that our method has the combined advantage of high fidelity, real-time rendering and consistency. GTM achieved state-of-the-art reconstruction quality. Rendering can be done at 80FPS with a graphics card, 100 times faster than NeRF-based methods. GTM also enable smooth appearance interpolation since it disentangles appearance from geometry.

In summary: \begin{itemize}
    \item We propose Guassian Time Machine(GTM) to reconstruct 3D scenes with discontinuous appearance variations. To our knowledge, it is among the first attempts to model long-term variations in real-time. Extensive experiments on challenging real-world datasets demonstrate that GTM outperforms state-of-the-art novel view synthesis methods for dynamic scenes in both rendering quality and efficiency, making it viable for real-time applications. 
    \item We validated that GTM disentangles appearance variations from geometry so that a smooth transition between different appearances can be achieved with GTM. 
\end{itemize}

\section{Related Works}
\subsection{Novel View Synthesis}
Novel view synthesis is the problem of predicting what a 3D scene is like from unseen viewpoints. Light-Field and Multi-View-Stereo based methods\cite{llff, Yao_2018_ECCV, PyramidMVS} are effective approaches to novel view synthesis. Recently, neural volume representations have achieved state-of-the-art rendering quality, among which NeRF is a popular one. NeRF\cite{mildenhall2020nerf} models a 3D scene with a field function (represented by an MLP) that maps spatial coordinates and view directions to color and density. Novel views are rendered in a volume rendering manner by sampling the function and accumulating the colors according to their densities. Given camera pose and ground truth image pairs, the implicit function is learned by minimizing the difference between the rendered image and the ground truth image. Vanilla NeRF boasts photo-realistic rendering quality but suffers from high computation cost and also requires dense input views. Therefore, several works\cite{kurz-adanerf2022, SunSC22, mueller2022instant, Chen2022ECCV, Reiser2023SIGGRAPH, kplanes_2023, duckworth2023smerf} have proposed various improved feature processing and sampling strategies and achieved more efficient training and rendering. Other studies\cite{pan2022activenerf, GeoAug} have enhanced the robustness of NeRF by uncertainty quantification or data augmentation.

3D Gaussian splatting\cite{kerbl20233d}, however, takes another approach to improve efficiency. It replaced the implicit field with a point-based explicit representation of 3D structures. Thus neural rendering can be accomplished in a rasterization manner, without repetitive sampling and querying neural networks. 3DGS enables fast training and real-time rendering for large-scale complex scenes, bringing great potential to further developments and applications. Its variants \cite{lu2023scaffoldgs, franke2024trips} enhance the representation with neural features, making it organized and more robust to view changes.

\subsection{Dynamic View Synthesis}
To adapt neural rendering to dynamic scenes, several methods have been proposed. Immersive Light Field\cite{broxton2020immersive} is built among videos recorded by 46 cameras with multi-spherical planes and can recover photorealistic videos in covered angles. Efficient-NeRF\cite{lin2022efficient} takes use of geometric consistency on videos recorded in different angles and guides ray sampling with predicted depth, enabling interactive free-viewpoint videos. For monocular videos, D-NeRF\cite{pumarola2021d} applies deformation fields to link alteration to the canonical scenes. On the other hand, some recent approaches \cite{Li_2023_CVPR,gao2021dynamic, nerfplayer, dfrf, hypernerf} take time t as the input spatial coordinate of the model to implement the space-time radiance fields. Due to the extra time dimension, these methods all fall short of real-time rendering. Thanks to the efficient rasterization-like training and rendering of 3DGS, several recent works\cite{4dgs, chen2023periodic, li2023spacetime, chen2023periodic, yan2023streetgaussians} have achieved real-time high-quality rendering on real-world dynamic scenes. 

Whether from multi-view or monocular videos, the aforementioned approaches assume space-time continuity of the training data, which does not hold for scenes captured at long intervals with complicated appearance variances.

\subsection{Neural Rendering with Complicated Varying Appearance}
A more challenging task is to reconstruct the scenes from images that are taken at discrete time intervals, which we try to address in this paper. Different from dynamic scenes that exhibit seamless continuity across space and time, under the discrete scenario, the variations of the scene are not continuous.  % 添加relighting-based appearance modeling

There are mainly two types of solutions to such a scenario. One line of approaches\cite{osr, Wang_2023_CVPR, nerfactor, PS-NeRF} decomposes the scene representation into rendering properties, including normal, base color, and materials, and predicts a set of parameters to relight the scene. Such physical-based-rendering methods are more interpretable and allow more precise control, but introduce extra cost. Another line of approach models such variances with neural encoders. NeRF-W\cite{Martin-Brualla_2021_CVPR} incorporates appearance embedding to the NeRF framework and deals with foreground occluders with the combination of a transient head and a predicted uncertainty level. With NeRF-W, scenes with controllable appearance can be reconstructed from unstructured internet photo collections of famous landmarks involving many images under uncontrolled settings. Block-NeRF and UC-NeRF\cite{tancik2022block, cheng2024ucnerf} reconstruct large outdoor scenes captured by automobiles and they all utilized latent appearance code similar to NeRF-W to increase the robustness of the model because the data was captured under varying environmental and lighting conditions. CLNeRF\cite{Cai_2023_ICCV} introduces continual learning to NeRF so that appearance and geometry changes across time could be updated with a limited number of new inputs given the prior knowledge from NeRF built upon with the old data. To our knowledge, there is no existing approach applying such a neural time encoder to 3DGS representation.

\section{Preliminaries}
\subsection{3D Gaussian Splatting (3DGS)} \label{3dgs}
3D Gaussian Splatting (3DGS) is a neural rendering representation. A scene is represented by a set of discrete Gaussian primitives with 5 parameters: \begin{center}
    $\mathbf{\Phi} = \left\{ \mathbf{\mu}_k, o_k, \mathbf{s}_k, \mathbf{r}_k, \mathbf{SH}_k \right\}_{k=1}^N$,
\end{center} where $\mu_k \in \mathbb{R}^3$ stands for the spatial center, $o_k \in \mathbb{R}$ for opacity, $\mathbf{s}_k \in \mathbb{R}^3$ for scaling, $\mathbf{r}_k \in \mathbb{R}^4$ for rotation quaternion, $\mathbf{SH}_k$ for spherical harmonics function coefficients that represents view-dependent colors, whose dimension can be adjusted according to the degree. Each primitive represents a Gaussian distribution that specifies spatial density: \begin{center}
    $\mathbf{p}(\mathbf{x}) = \exp({-\frac{1}{2}(\mathbf{x} - \mu)^T\mathbf{\Sigma}^{-1}(\mathbf{x} - \mu)})$
\end{center}, where the covariance term is calculated as $\mathbf{\Sigma} = RSS^TS$ so as to keep it positive-definite.

The primitives are rendered in a process similar to rasterization. Given the pose of a camera, they are first projected to screen space by the viewing matrix W and projection matrix J: $\mathbf{\Sigma}_{\textnormal{2d}} = JW\mathbf{\Sigma}W^TJ^T$, with those outside the view frustum filtered out. Then the view-dependent color is obtained by querying the Spherical Harmonics function with the camera pose: $\mathbf{c}_k = \textnormal{SH}_k(P)$. Finally, screenspace colors are blended according to the cumulative density: \begin{center}
    $\mathbf{c} = \mathop{\sum}\limits_{i=1}^N T_io_i\mathbf{c}_i$, \hspace{0.5cm} $T_i = \mathop{\prod}\limits_{j=1}^{i-1} (1-o_j)$
\end{center}

The whole rendering process is differentiable and can be described using a function: $\hat{\mathcal{I}} = \mathcal{R}(\mathbf{P}; \mathbf{\Phi})$ The training process is similar to NeRF: given a training set with pairs of image and corresponding camera pose $\mathcal{S} = \left\{\mathcal{I}, \mathbf{P}\right\}$,  the parameters are optimized using gradient descent to minimize the error between the rendered image and the ground truth.

The number of Gaussian primitives is adjusted during training. Gaussian primitives are periodically densified. Primitives with large gradients are split into two and those with small opacity are pruned. Such a strategy is the key to 3DGS's success in reconstructing fine-grained details.

\subsection{Scaffold-GS} \label{scaffold-gs}
A shortcoming of the densification process is that such a strategy tends to produce redundant Gaussian points and unsatisfactory effects on texture-less regions. To address these limitations, Scaffold-GS\cite{lu2023scaffoldgs} is proposed. It organizes Gaussian primitives in a $k$-tree structure to avoid over-densification. A set of learnable anchor points $\mu_a^{(i)}$ along with corresponding feature vectors $f$, offset vectors $\mathbf{v}_o$ and scaling vectors $\mathbf{s}$ \begin{center}
    $\left\{\mu_a^{(i)}, f^{(i)}, \mathbf{v}_o^{(i)}, \mathbf{s}^{(i)}\right\}_{i=1}^N$
\end{center}  are first initialized, and k neural Gaussians are generated for each anchor point. 

Each neural Gaussian primitive has five properties as in 3DGS: $\mathbf{\mu}, o, \mathbf{s}, \mathbf{r}, \mathbf{c}$. The centers of neural Gaussians are calculated by adding offset vectors to the center of the anchor point: \begin{center}
    $\mathbf{\mu} = \mu_a + \mathbf{v}_o \cdot \mathbf{s}$
\end{center}
where $\mathbf{v}_o \in \mathbb{R}^{k\times 3}$ is the offset vector, $\mathbf{s} \in \mathbb{R}^3$ is the scaling vector, $\cdot$ denotes matrix multiplication, $\mu_a \in \mathbb{R}^3$ denotes the center of the anchor point, these two terms are added with broadcast and  $\mu \in \mathbb{R}^{k\times 3}$ is the compressed neural Gaussian centers. The vectors $\mathbf{v}_o$ and $\mathbf{s}$ are directly optimized.

The rest of the properties are predicted by lightweight Multi-Layer Perceptrons (MLPs) with activation functions:  \begin{center}
    $\mathbf{o} = \tanh({F_o(f, \delta_{vc},\mathbf{d}_{vc})})$, \\
    $\mathbf{c} = \sigma(F_c(f, \delta_{vc},\mathbf{d}_{vc}))$, \\
    $(\mathbf{s, r}) = F_{cov}(f, \delta_{vc},\mathbf{d}_{vc})$
\end{center}    
where $f$ is a feature corresponding to an anchor point, $\delta_{vc}, \mathbf{d}_{vc}$ are the distance and relative direction between the camera and the anchor point, $\sigma(\cdot)$ denotes the Sigmoid activation function. Here the scaling and rotation properties are decoded by one single network and split into each. 

It is worth noting that view-dependent effects are represented by spherical harmonics function in 3DGS, but in Scaffold-GS, the color is inherently view-dependent because view vectors are included in the input of the color MLP.

\section{Method}
In this section, we will first formulate the problem. Then key designs that enable Gaussian Time Machine to model long-term environmental changes will be presented. 
\subsection{Problem Formulation}
Gaussian Time Machine is proposed for novel view synthesis on scenes with varying appearances. Given a training set with images of the same place or objects taken at different times $I = \left\{(\mathcal{I}_k, \mathbf{P}_k, t_k )\right\}_{k=1}^N$, where $I_k$ denotes the image, $\mathbf{P}_k$ denotes the camera parameters (extrinsic matrix, intrinsic matrix and canvas size), $t_k$ denotes the time. The goal is to synthesize novel views with combinations of $(\mathbf{P}, t)$ that are not in the training set, which is achieved by learning the parameters of a function that maps from image pose and time step to RGB values: $\hat{I}_k = F(\mathbf{P}_k, t_k; \mathbf{\Phi})$.

\begin{figure}
    \centering
    \includegraphics[width=\linewidth, trim = {0, 4cm, 0, 5cm}, clip]{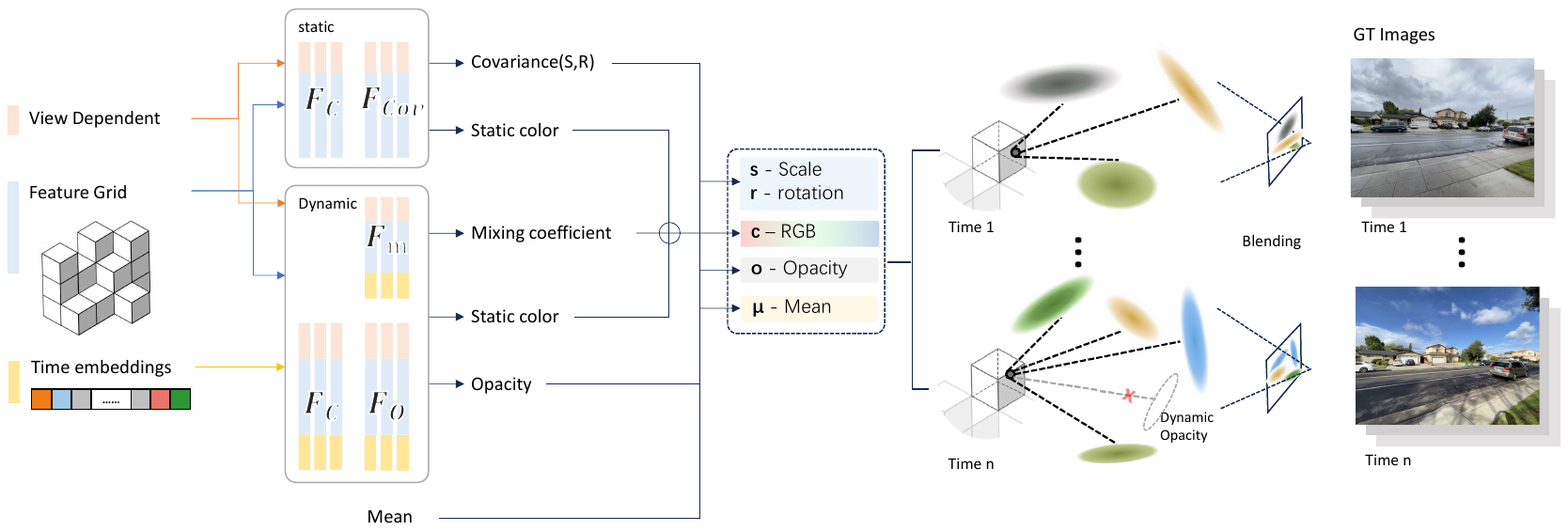}
    \caption{Overview of Gaussian Time Machine(GTM). A set of lightweight neural networks take discrete time embedding vectors as input, predicting the time-variant properties of neural Gaussian primitives. Then neural Gaussians are blended to render images. The network parameters and embedding vectors are jointly optimized by minimizing the loss function between the predicted images and ground-truth images, along with a regularization term.}
    \label{fig:overview}
\end{figure}

\subsection{Modeling Varying Gaussians}
Visibility changes are common in time-variant scenes that can't be modeled with a fixed number of Gaussian primitives over the time range. To address this challenge, it's necessary modify the number of Gaussian primitives on the fly to adapt the model to complex scene variations. 

One way to implement such a design is to simultaneously predict the number and properties of changing Gaussian primitives with a single neural network, which is challenging for training. Therefore, we take another approach: generating a sufficiently large number of Gaussian primitives first, then adjusting their visibility. Scaffold-GS encoding network\cite{lu2023scaffoldgs} is utilized as the architecture of the encoder in GTM, whose k-tree structure generates multiple Gaussian primitives around an anchor point in an organized way, striking a balance between rendering details and efficiency. The opacity term serves as the key to adjusting the number of Gaussians. It is activated with a hyperbolic tangent function: $\tanh{(\cdot)} \in [-1, 1]$. The Gaussian primitives with negative activation values are filtered out when being rendered. Therefore, if the visibility of a specific Gaussian primitive changes over time, then the MLP can be optimized to predict its activation value as negative at some time steps but positive at other time steps. Our approach for modeling a varying number of Gaussian primitives is illustrated in the right half of Fig. \ref{fig:overview}.

\subsection{Time Step Encoding Scheme} \label{time_step_encoding}
Integrating time-dependent information into a 3D scene is the key to modeling long-term variances. In 3DGS-based neural rendering, the five properties of Gaussian primitives time-dependent should be adjusted using an encoder. There are mainly 3 types of time encoder: probabilistic, continuous mapping, and discrete embedding. 

The probabilistic encoder estimates a joint distribution of primitive parameters $p(\mathbf{\Phi},t)$. 4DGS\cite{4dgs} approximated the joint distribution by reparameterizing each domain of Gaussian primitives and jointly optimizing the 4D distribution. 

The continuous mapping encoder utilizes lightweight neural networks to model the correlation between space and time. Time step inputs are expanded to higher dimensions and fed into the neural network. The network then predicts the time-dependent parameters. In Deformable-3DGS, the mapping function is the Position Encoding function $\gamma(t) = (\sin(2^k\pi{t}), cos(2^k\pi{t}))_{k=0}^{L-1}$, which acts as an enhancement of features.

These two approaches all assume that the training data is a continuous space-time sequence with minor inter-frame differences, which does not hold for the case that we are dealing with. Moreover, continuous encoders tend to overfit on the discrete input time steps, resulting in broken geometries at intermediate time steps (see Fig. \ref{fig:interpolation}). The discrete embedding encoder, contrary to continuous encoders, models the changing properties of Gaussian primitives with a set of learnable vectors and optimizes them together with the MLP. It is most suitable for modeling long-term appearance variances. In GTM, the learnable embedding vector\cite{optimizing-latent-space} is concatenated with the input to each MLP encoder to model time-dependent variances. The properties of neural Gaussians are calculated as:\begin{center}
 $\mathbf{z}_t = \textnormal{TimeEmbedder}(t)$, \\
 $\mathbf{o} = \tanh(F_o(f, \delta_{vc},\mathbf{d}_{vc}, \mathbf{z}_t))$, \\
 $\mathbf{c} = \sigma(F_c(f, \delta_{vc},\mathbf{d}_{vc}, \mathbf{z}_t))$, \\
 $(\mathbf{s, r}) = F_{cov}(f, \delta_{vc},\mathbf{d}_{vc})$
\end{center}
where $\mathbf{o} \in \mathbb{R}^{k\times 1}$ and $\mathbf{c} \in \mathbb{R}^{k\times3}$ are the opacities and RGB colors of $k$ neural Gaussians, $\mathbf{z}_t \in \mathbb{R}^l$ is the time embedding vector, and the rest variables follow the definition in \ref{scaffold-gs}. Note that the rotation and scaling (Gaussian covariance) terms are computed as time-independent to improve geometry consistency. 

 \subsection{Decomposed Color Blending}
 To better reconstruct scene dynamics, color is rendered in a decomposed manner. To formulate the decomposition, we refer to color formation theories in physics-based rendering. When performing ray tracing, the color of an object is comprised of the color of the object itself and reflection from the environment. In scenes with long-term variations, usually, it is the lighting or weather that varies over time, while the objects remain the same. 
 
 Inspired by this light formation model, we divide the color into a static term and a dynamic term. The static term that remains constant over time steps is computed by a time-invariant MLP. The dynamic MLP, on the contrary, integrates time embedding as input to model varying environment lighting: \begin{center}
     $\mathbf{c}_s = \sigma(F_{\mathbf{c}_s}(f, \delta_{vc},\mathbf{d}_{vc}))$, \\
     $\mathbf{c}_d = \sigma(F_{\mathbf{c}_d}(f, \delta_{vc},\mathbf{d}_{vc}, \mathbf{z}_t))$
 \end{center}
 The definition of the input vectors to the network is defined in Section \ref{time_step_encoding}. In the color formation model, the material of objects affects how much the reflected environment lighting accounts for the rendering color. For instance, a metal surface displays mostly the color of its surroundings while a wooden surface displays mostly its own color. To model such an effect, we use another MLP to estimate a blending coefficient that adaptively blends these two terms: \begin{center}
     $m = \sigma(F_m(f, \delta_{vc},\mathbf{d}_{vc}, \mathbf{z}_t))$
 \end{center}
 The composite color is computed as \begin{center}
     $\mathbf{c} = (1 - m)\mathbf{c}_s + m\mathbf{c}_d$
 \end{center}
 Finally, the colors of neural Gaussian primitives are blended according to weights computed by accumulated opacity as described in Section \ref{3dgs}.

 \subsection{Optimization}
 3DGS is trained by solving the optimization problem using gradient descent: \begin{center}
     $\mathbf{\Phi}^* = \mathop{\textnormal{argmin}}\limits_{\mathbf{\Phi}} \mathcal{L}(\mathcal{R}(\mathbf{P}; \mathbf{\Phi}))$
 \end{center} 
 In GTM, time-dependent information is integrated and the parameters of Gaussian primitives $\mathbf{\Phi}$ are predicted by a neural encoder instead of being directly optimized. The optimization goal is reformulated as: \begin{center}
     $\mathbf{\Phi} = \mathbf{F}_{\mathbf{\Theta}}(f, \delta_{vc}, \mathbf{d}_{vc}, \mathbf{z}_t)$, 
     $\mathbf{\Theta}^* = \mathop{\textnormal{argmin}}\limits_{\mathbf{\Theta}} \mathcal{L}(\mathcal{R}(\mathbf{P}, t; \mathbf{\Phi}))$,
 \end{center}
 where the combined parameter $\mathbf{\Theta}$ includes the parameters of encoder networks and learnable time embedding vectors. The loss function \begin{center}
     $\mathcal{L}(\hat{\mathcal{I}}) = ||\hat{\mathbf{\mathcal{I}}} - \mathbf{\mathcal{I}}||_1 + \lambda_{SSIM}\textnormal{SSIM}(\hat{\mathbf{\mathcal{I}}}, \mathbf{\mathcal{I}}) + \lambda_{\textnormal{vol}}\mathcal{L}_{\textnormal{vol}}$
 \end{center},
 is a combination of the volumetric loss with the pixel-wise L1 image loss, SSIM loss proposed by 3DGS, and the volume regularizing term $\mathcal{L}_{\textnormal{vol}}$, which is calculated by summing up the product of scaling vector on all dimension: \begin{center}
     $\mathcal{L}_{\textnormal{vol}} = \mathop{\sum}\limits_{i=1}^N \textnormal{Prod}(\mathbf{s}_i)$
 \end{center}
 where $N$ denotes the number of neural Gaussian primitives and $\textnormal{Prod}(\cdot)$ denotes the product of a vector on all of its dimensions.

\section{Experiments}

\subsection{Experimental Setup}
\subsubsection{Dataset} GTM is tested on 3 real-world public datasets: World Across Time(WAT), Phototourism, and NeRF-OSR. All of these datasets contain images taken at different seasons or under different weather conditions. WAT and Phototourism include both camera poses and point clouds estimated using COLMAP\cite{Schonberger_2016_CVPR}, a structure-from-motion (SfM) software. NeRF-OSR, however, estimated camera poses in the same manner but provided camera poses without SfM point cloud. Since our method requires sparse SfM point clouds for initializing anchor points\cite{kerbl20233d}, the dataset is reprocessed using COLMAP.

\subsubsection{Baselines} We choose the following studies as our baselines: 1) NeRF in the wild (NeRF-W), a variant of NeRF that reconstructs the static central object from photos captured with occlusion and varying light conditions. 2) CLNeRF, a continual learning framework for time-varying scenes, which also proposed the WAT benchmark. 3) Deformable-3DGS, an extension to 3DGS that calculates the movements of Gaussian primitives using a deform field and a lightweight MLP, assuming that the number of Gaussians does not vary over time. It performs well in reconstructing dynamic scenes from video snapshots. 4) 4DGS, an extension of 3DGS to dynamic scenes that integrates the dynamics of Gaussian primitives into the differentiable Gaussian rendering function by predicting a 4D joint distribution of time-variant Gaussian primitives.

\subsubsection{Evaluation Methodology} On WAT and Phototourism, all the images are rendered at their original resolution. On NeRF-OSR, renderings are 4x downsampled (1280p). For the quantitative study, PSNR, SSIM, and LPIPS compared with ground truth photos are chosen as evaluation metrics for rendering quality. On Phototourism, we use the same training-test split as in NeRF-W\cite{Martin-Brualla_2021_CVPR}. For evaluation of rendering speed, the inference time on a single NVIDIA RTX 3090 GPU is reported in FPS. Moreover, the average storage space is measured since it will influence the efficiency of model deployment. 

\begin{figure}   
    \centering
    \includegraphics[width=\linewidth, trim={6.4cm, 0cm, 6.4cm, 0.4cm}, clip]{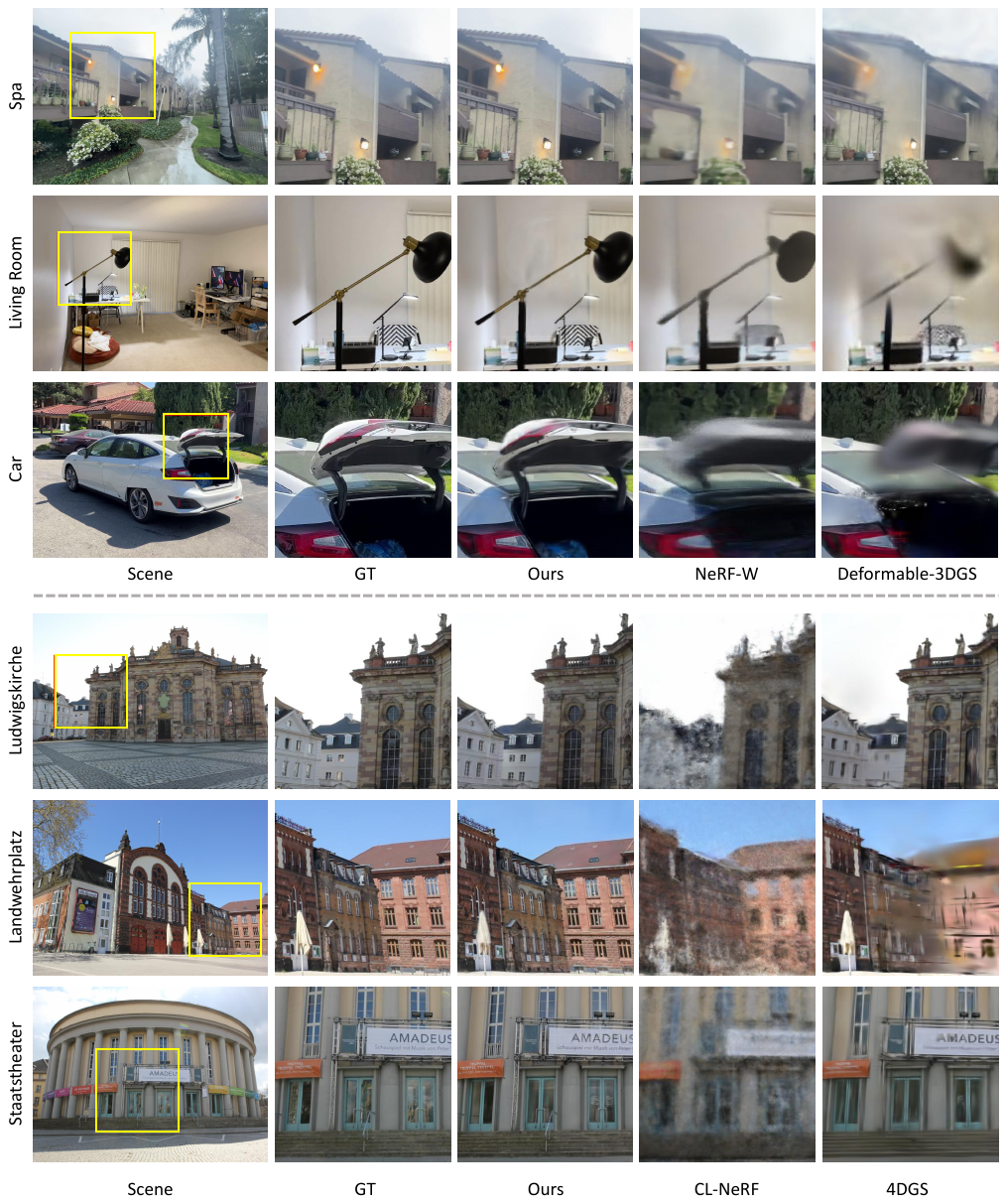}
    \caption{Visual comparison of rendering quality between GTM and existing methods on WAT\cite{Cai_2023_ICCV} dataset (first 3 rows) and NeRF-OSR\cite{osr} dataset (the rest). NeRF-based approaches with time embeddings can disentangle appearance changes but limited representation capability often leads to floater artifacts. Continuous-time 3DGS methods have stronger representation capability, but due to the limitation of their time encoding scheme, they tend to overfit training images, or cannot reconstruct complex variations, leading to blurry artifacts. GTM combines the strengths of the two approaches and can produce high-fidelity renderings, especially on details and distant parts of the scene. }
    \label{fig:visual comparison}
\end{figure}

\subsection{Quantitative Study Results}

\begin{table}[htbp]
    \centering
    {\fontsize{6}{8}\selectfont
    \begin{tabular}{c|c|ccc|ccc|ccc|cc}
    \toprule
      \multicolumn{2}{c|}{Dataset} & \multicolumn{3}{c|}{WAT} & \multicolumn{3}{c|}{Phototourism} & \multicolumn{3}{c|}{NeRF-OSR} & \multirow{2}{*}{FPS$\uparrow$} & \multirow{2}{*}{Disk(MB) $\downarrow$}\\
      Method & Metrics & PSNR$\uparrow$ & SSIM$\uparrow$ & LPIPS$\downarrow$ & PSNR$\uparrow$ & SSIM$\uparrow$ & LPIPS$\downarrow$ & PSNR$\uparrow$ & SSIM$\uparrow$ & LPIPS$\downarrow$ & ~ & ~ \\
      \hline
      \multicolumn{2}{c|}{NeRF-W} & 25.03 & 0.759 & 0.350 & \cellcolor{third} 23.13 & \cellcolor{third} 0.827 & \cellcolor{third} 0.227 & \cellcolor{third} 19.74 & 0.531 & 0.518 & 0.032 & \cellcolor{first} 16\\
      \multicolumn{2}{c|}{CL-NeRF(UB)} & 25.45 & 0.773 & 0.361 & \cellcolor{second} 23.98 & \cellcolor{second} 0.827 & \cellcolor{first} 0.205 & 18.27 & 0.509 & 0.595 & 0.65 & \cellcolor{second} 54\\
      \multicolumn{2}{c|}{Deformable-3DGS} & 24.58 & 0.802 & 0.309 & - & - & - & 19.14 & \cellcolor{third} 0.691 & \cellcolor{third} 0.364 & \cellcolor{first}{124.61} & 80\\
      \multicolumn{2}{c|}{4DGS} & \cellcolor{third} 25.74 & \cellcolor{third} 0.829 & \cellcolor{third} 0.288 & - & - & - & 18.74 & 0.637 & 0.438 & 79.84 & 3664\\ \hline
      \multicolumn{2}{c|}{{\bf GTM(ours)}} & \cellcolor{first} 27.62 &  \cellcolor{first} 0.873 &  \cellcolor{first} 0.216 & \cellcolor{first} 25.30 & \cellcolor{first} 0.854 & \cellcolor{second}{ 0.208} & \cellcolor{first} 21.27  &  \cellcolor{first} 0.741 & \cellcolor{first} 0.294 & \cellcolor{third} 80.71 & \cellcolor{third} 75\\
      \multicolumn{2}{c|}{PE time encoder} & \cellcolor{second} 27.38 & \cellcolor{second} 0.869 & \cellcolor{second} 0.225 & - & - & - & \cellcolor{second} 21.11 & \cellcolor{second} 0.736 & \cellcolor{second} 0.304 & \cellcolor{second} 88.53 & 93 \\
    \bottomrule
    \end{tabular}}
    \caption{Quantitative study results on rendering quality and efficiency between GTM and existing methods. We reproduced all the baseline methods on 3 datasets using open-source implementation and reported the results. ('-' denotes that the method cannot reach stable convergence on the dataset.) GTM consistently achieves state-of-the-art performance in terms of rendering quality while maintaining real-time rendering speed. As for storage space, GTM uses the least disk space among 3DGS-based methods.}
    \label{tab:my_label}
\end{table}

Quantitative study shows that GTM has combined strengths of rendering quality and efficiency. Compared to NeRF-based methods that model long-term visual variations with embedding vectors, GTM can reconstruct fine-grained details because of the improved representation capability of Gaussian primitives. Compared with 3DGS-based methods for dynamic scenes, GTM can disentangle the appearance variation from geometry and control the number of Gaussian primitives, thus achieving state-of-the-art rendering quality. GTM is also much more efficient in terms of storage space, with the trainable latent code as a guiding feature to disentangle environment variations, thus avoiding over-reconstruction and requiring less storage space (less than 100MB per scene on average). 

Another version of GTM is also tested, in which the time embedding vector is replaced with a continuous encoder (Positional Encoding). The performance of discrete embedding vector encoder is slightly better on WAT and NeRF-OSR datasets, but the improvement on the Phototourism dataset is significant. Continuous time encoder fails to lead the training session to convergence when the space-time appearance variations are complicated. The discrete embedding vector encoder, boosts robustness and enables GTM to scale to a larger set of appearances.

\subsection{Qualitative Study Results}

Fig. \ref{fig:visual comparison} shows the results of the qualitative study. We visually compare the renderings from GTM and the baselines. NeRF-based methods can model long-term changes in lighting and environment and perform well on object-centric scenes. However, unbounded scenes can be challenging to reconstruct. 3DGS variants for dynamic scenes can reconstruct dynamic scenes with fine-grained details. However, when the dynamics of the scene are not continuous or the visibility of some objects changes, GTM is the more promising method. With 3D Gaussian as primitives of scene representation, accompanied with the time embedding, GTM can not only encode complex variations but also capture fine-grained details. We do not limit the number of visible Gaussian primitives to be constant but adjust their color and opacity on-the-fly. With these strategies, GTM generates less blurry artifacts.

\begin{figure}
    \centering
    \includegraphics[width=\linewidth, trim={0.8cm, 1cm, 1cm, 1cm}, clip]{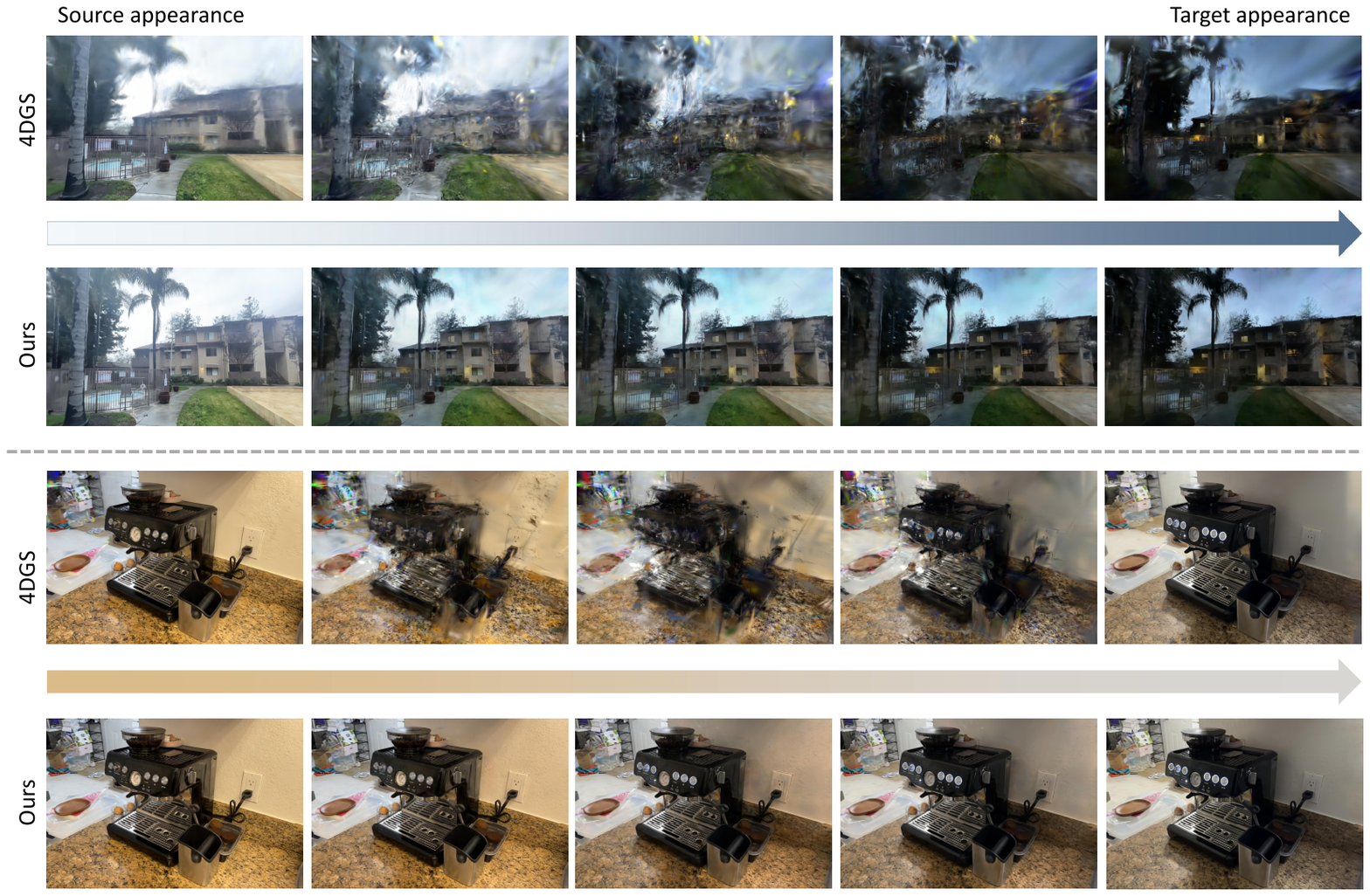}
    \caption{Visual comparison between 4DGS and GTM on appearance control. We select two scenes from WAT. For GTM, we linearly interpolate the embedding vector. For 4DGS, we train on discrete time stamps and assign a continuous sequence of time steps in between. Interpolation on GTM produces consistent geometry over time and the trajectories of most Gaussian primitives are ordered, so there are less artifacts. These results demonstrate that GTM's time encoding scheme can disentangle the variance of color from the invariant part of the scene.}
    \label{fig:interpolation}
\end{figure}

With GTM, the appearance can be controlled while maintaining consistent geometry. In Fig. \ref{fig:interpolation}, the time-dependent input is linearly interpolated and renderings from GTM and 4DGS are shown. Results show that linear interpolation on the time embedding vector $\mathbf{z}_t$ of GTM only changes the lighting, generating a smooth transition. GTM represents time-variant information in a discrete manner, so it will not overfit discrete training images or produce broken geometries. In general, GTM enables both a controllable appearance and a smooth transition from one style to another.

\subsubsection{Decomposition Visualization}
\begin{figure}
    \centering
    \includegraphics[width=\linewidth, trim={0.5cm, 8cm, 0.5cm, 8.2cm}, clip]{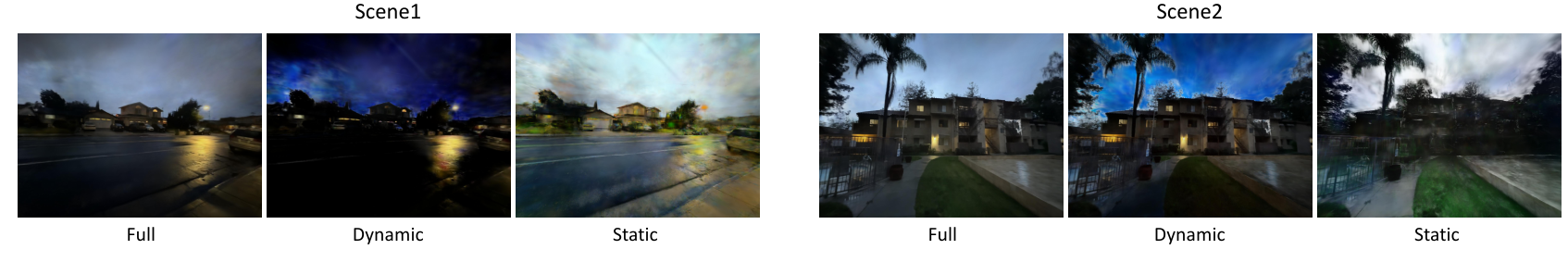}
    \caption{Visualization of the static head, dynamic head, and full rendering results. The static head mainly models the underlying geometry while the dynamic head captures the lighting.}
    \label{fig:head_visualization}
\end{figure}

To validate the effectiveness of the decomposition design of GTM, we rendered the static and dynamic terms respectively in Fig. \ref{fig:head_visualization}. Visualization results show that such decomposition achieved an understanding of scene components. The dynamic head mainly accounts for the varying lighting conditions while the static head models underlying geometry.

\begin{figure}
    \centering
    \includegraphics[width=\linewidth, trim={0, 5cm, 0, 5cm}, clip]{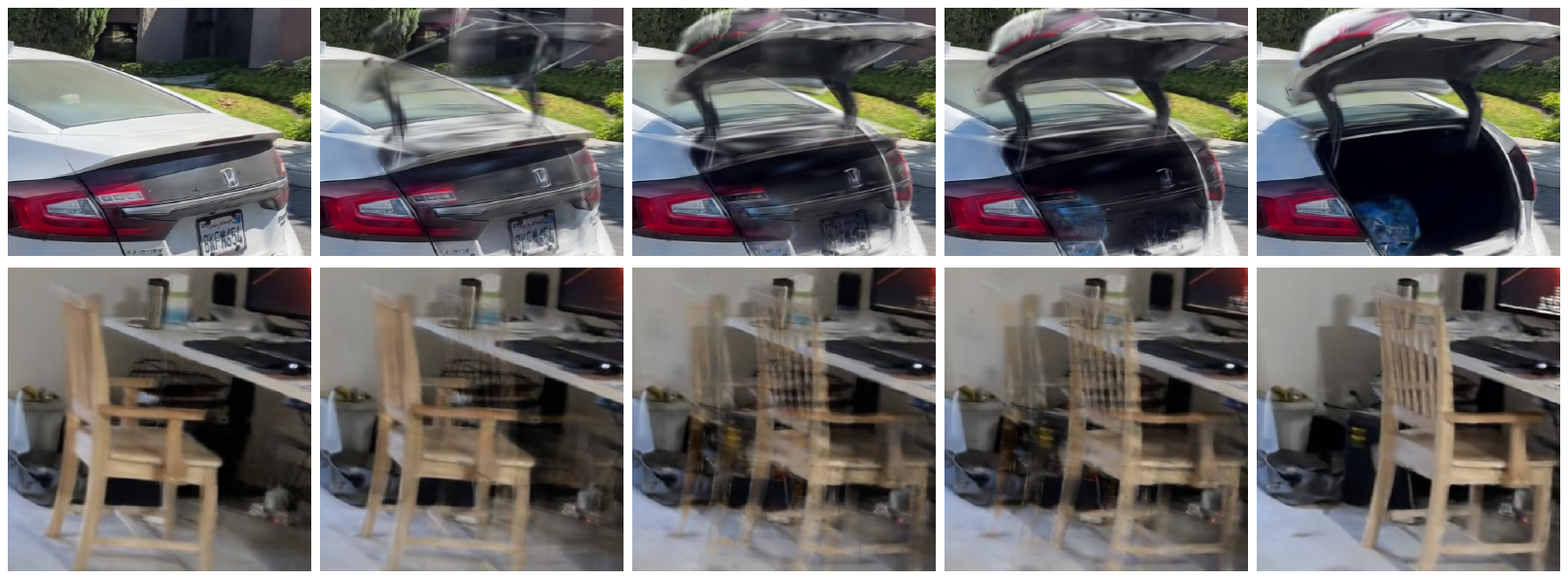}
    \caption{Cases where GTM reconstruction results are inconsistent with physical movement. Dynamic opacity can accurately control the visibility of the desired part of the scene, but the interpolation results are gradual fading-in and fading-out of specific parts, which is inconsistent with the true movement trajectory.}
    \label{fig:dynamic_opacity}
\end{figure}

\section{Conclusion}
We presented Gaussian Time Machine (GTM), a method for novel view synthesis on scenes with time-variant appearances. We successfully addressed the challenges of visibility changes and discontinuous variations. The Scaffold-GS encoding network is adopted and enhanced with a time embedding vector that encodes the time-variant properties of neural Gaussians. By adjusting the opacity of neural Gaussians on the fly, GTM can capture fine-grained appearance changes. The color prediction head is decomposed into a dynamic and static one, improving understanding of the scene. Gaussian Time Machine has enabled fast training and real-time rendering of scenes with complex variances over long, discontinuous time span, outperforming both NeRF-based and 3DGS-based counterparts on challenging real-world datasets. Moreover, our appearance model is controllable and a smooth appearance transition between different styles can be achieved by interpolating embedding vectors. % flexibility on modeling both 

With a combination of fast speed, high rendering fidelity, and low training cost, GTM has the potential to be used in various visualization applications. Existing street view maps have a 'time machine' function that enables users to wander through city spaces over time, but only at fixed viewpoints because it adopts panorama as the means of 3D visualization. GTM can enhance street view applications, enabling 6DOF free viewpoint roaming. GTM can also be applied in digital twins, which requires interaction with the model, as in the case of opening the door or trunk lid of a reconstructed virtual car depicted in Fig.\ref{fig:teaser}.

Despite the capability of high-quality and real-time rendering of scenes with complex appearance variations, GTM can still be improved in terms of scene understanding. GTM deals with the changing visibility of objects by dynamic opacity control, which is an approximation and is inconsistent with physical laws. Understanding physical movement and producing physically consistent renderings within varying appearances is left for future work.


\bibliographystyle{IEEEtran}
\bibliography{main}
\end{document}